\documentclass[11pt]{article}

\usepackage[final]{acl}

\usepackage{times}
\usepackage{latexsym}
\usepackage{booktabs}
\usepackage{makecell}
\usepackage{tabularx}
\usepackage{booktabs}
\usepackage[T1]{fontenc}

\usepackage{xcolor}
\usepackage[most]{tcolorbox}
\usepackage{listings}

\usepackage{fvextra}

\usepackage[most]{tcolorbox}
\usepackage{xcolor}

\usepackage[utf8]{inputenc}
\usepackage{pdfpages}
\usepackage{microtype}

\usepackage{inconsolata}

\usepackage{graphicx}
\newcommand{\hide}[1]{}

\newcommand{\AHAMZAUP}{{\^{A}}}

\newcommand{\TAMARBUTA}{{$\hbar$}}

\newcommand{\SHIN}{{\v{s}}}
\newcommand{\ZA}{{\v{D}}} 
\newcommand{\AYN}{{$\varsigma$}}

\newcommand{\KASRATAN}{{\~{\i}}}

\newcommand{\SHADDA}{{$\sim$}}

\usepackage{cleveref}
\usepackage{booktabs}
\usepackage{graphicx}
\usepackage{multirow}

\usepackage{microtype}
\usepackage{siunitx}
\usepackage{todonotes}
\usepackage{arydshln}

\usepackage{enumitem}
\usepackage[super]{nth}
\usepackage{tipa}
\usepackage{svg}
\usepackage{adjustbox}

\usepackage{stackengine}
\usepackage{tikz}
\newcommand\dottedcircle{\tikz \draw [line cap=round, line width=0.15ex, dash pattern=on 0pt off 1.95\pgflinewidth] (0,0) circle [radius=0.6ex];}
\makeatletter

\newcommand{\yallamorph}{\bf YallaMorph}

\newcommand{\fatha}{%
\bgroup \set@arabfont
\stackon[0.5pt]{\dottedcircle}{\char'013}%
\egroup%
}
\newcommand{\kasra}{%
\bgroup \set@arabfont
\stackunder[0.5pt]{\dottedcircle}{\char'013}%
\egroup%
}
\newcommand{\damma}{%
\bgroup \set@arabfont
\stackon[0.5pt]{\dottedcircle}{\char'014}%
\egroup%
}
\newcommand{\fathatan}{%
\bgroup \set@arabfont
\stackon[0.5pt]{\dottedcircle}{\char'023}%
\egroup%
}
\newcommand{\kasratan}{%
\bgroup \set@arabfont
\stackunder[0.5pt]{\dottedcircle}{\char'023}%
\egroup%
}
\newcommand{\shaddah}{%
\bgroup \set@arabfont
\stackon[0.5pt]{\dottedcircle}{\char'017}%
\egroup%
}
\newcommand{\dammatan}{%
\bgroup \set@arabfont
\stackon[0.5pt]{\dottedcircle}{\char'024}%
\egroup%
}
\newcommand{\skun}{%
\bgroup \set@arabfont
\stackon[0.5pt]{\dottedcircle}{\char'025}%
\egroup%
}
\newcommand{\alif}{%
\bgroup \set@arabfont
\stackon[0.5pt]{\dottedcircle}{\char'027}%
\egroup%
}
\makeatother

\usepackage{arabtex}
\usepackage{utf8}
\setcode{utf8}
 
\usepackage{epstopdf}

\title{{\yallamorph}: A Benchmark for Evaluating\\ Arabic Morphological Generation in Large Language Models}

\newcommand*{\authormark}[1][*]{\textsuperscript{#1}}
\author{
    Mahmoud Reda,\authormark[1]
    Salam Khalifa,\authormark[1,2]
    Reham Marzouk,\authormark[3]
    Nizar Habash\authormark[1]\\
    {\normalfont Computational Approaches to Modeling Language (CAMeL) Lab} \\
    {\normalfont\authormark[1]New York University Abu Dhabi,}\\
    {\normalfont\authormark[2]Stony Brook University,}
    {\normalfont\authormark[3]Mohamed bin Zayed University of Artificial Intelligence}\\
    {\normalfont\texttt{\{mahmoud.ali,salam.khalifa,nizar.habash\}@nyu.edu}, \texttt{Reham.Marzouk@mbzuai.ac.ae}}\\
}

\begin{document}
\maketitle


\begin{abstract}
Arabic morphology remains challenging for large language models, since fluent generation does not guarantee accurate morphosyntactic control. Existing Arabic evaluations mainly target downstream tasks and do not directly test controlled morphological generation from explicit lexical and feature-based input. We introduce {\yallamorph}, a large-scale benchmark for Arabic morphological generation covering verbs, nouns, adjectives, their cliticized forms, and invalid configurations. We evaluate multilingual and Arabic-oriented LLMs under diacritized and undiacritized settings over 600K benchmark entries. Results show that Arabic morphological generation remains difficult, especially for cliticized, unseen, and morphologically rare forms.
\end{abstract}

\section{Introduction}


Arabic morphology presents a major challenge for language generation.
The interaction of templatic and concatenative morphology with morphosyntactic and cliticization features creates a large space of productive forms. Our central question is not simply whether Large Language Models (LLMs) can generate plausible Arabic, but whether they can systematically produce appropriate inflected forms for specific morphological features.

Most evaluations of Arabic LLMs focus on downstream tasks or broad generation quality \cite{nagoudi-etal-2022-arat5,almazrouei-etal-2023-alghafa,koto-etal-2024-arabicmmlu}, providing limited insight into whether models can systematically generate correct Arabic word forms from explicit lexical and morphological specifications. Existing morphological inflection benchmarks provide more controlled evaluation settings \cite{kodner-etal-2022-sigmorphon,goldman-etal-2023-sigmorphon}, but they do not capture the scale and Arabic-specific feature space required for evaluating modern LLMs.

\begin{figure}[t!]
\centering
\includegraphics[width=\linewidth]{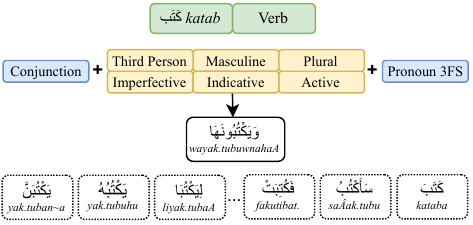}

 \caption{Morphological realization of Arabic lemma \<كَتَب> \textit{katab} `write' as \<وَيَكْتُبُونَهَا> \textit{wayak.tubuwnahaA} `and they write it' through morphosyntactic features and clitics. Transliterations follow HSB \citep{Habash:2007:arabic-transliteration}.}
\label{fig:task-overview}
\end{figure}

We introduce {\yallamorph}, a large-scale benchmark for controlled Arabic morphological generation.\footnote{\label{fn:yallamorph-repo}
The benchmark data is publicly available:
\url{https://github.com/CAMeL-Lab/YallaMorph}} As illustrated in Figure~\ref{fig:task-overview},
the task maps a lemma, part-of-speech, gloss, and target feature bundle to the corresponding Arabic surface form, or to an indication that the requested configuration is morphologically invalid. The benchmark covers verbs, nouns, adjectives, and their cliticized forms, and contains over 600K evaluation instances sampled across root class, stem complexity, paradigm completeness, and lemma frequency.

We evaluate proprietary, multilingual, and Arabic-oriented instruction-tuned LLMs under both diacritized and undiacritized settings. Results show that controlled Arabic morphological generation remains challenging even for strong modern LLMs, particularly for cliticized, unseen, and morphologically rare forms.

Our contributions are: (a) introducing a \textbf{large-scale benchmark} for controlled Arabic morphological generation designed using a linguistically motivated sampling framework; and (b) \textbf{evaluating multilingual and Arabic-oriented LLMs} and providing detailed analysis across linguistic and distributional dimensions.

\newpage
    
\section{Related Work}

\paragraph{Morphological Inflection}
Morphological inflection is a well-established standalone task in NLP. The SIGMORPHON morphological (re)inflection shared tasks have served as a central benchmark series for this problem \cite{cotterell-etal-2016-sigmorphon,cotterell-etal-2017-conll,cotterell-etal-2018-conll,mccarthy-etal-2019-sigmorphon,vylomova-etal-2020-sigmorphon,pimentel-ryskina-etal-2021-sigmorphon,kodner-etal-2022-sigmorphon,goldman-etal-2023-sigmorphon}. These tasks evaluate systems that map lemmas and morphosyntactic feature bundles to inflected forms, and later editions increasingly emphasized generalization across typologically diverse languages, unseen lemmas, and unseen feature combinations. The 2022 and 2023 shared tasks in particular strengthened the evaluation setup through data splits targeting generalization to unseen lemmas \cite{kodner-etal-2022-sigmorphon,goldman-etal-2023-sigmorphon}. This line of work is closely tied to UniMorph \cite{batsuren-etal-2022-unimorph}, a broad-coverage universal schema for morphological annotation and inflection tables that represent forms through a lemma and a bundle of morphosyntactic features. UniMorph 4.0 covers 182 languages, including Arabic varieties such as Modern Standard Arabic (MSA), Egyptian Arabic, and Gulf Arabic in recent shared-task data. However, while the SIGMORPHON shared tasks yielded a diverse set of baselines and state-of-the-art systems, they were not designed as LLM benchmarks. Moreover, for our purposes, UniMorph does not provide exhaustive lexical coverage, full Arabic paradigm coverage, or cliticized inflected forms. In this work, we instead use Arabic-specific morphological resources that provide broader lexical coverage and richer paradigm generation.

\paragraph{Morphological Generators for Arabic}
Morphological analyzers and generators have been central to Arabic NLP since its early stages, providing explicit linguistic representations for a morphologically rich and complex language. Early work included finite-state and templatic approaches that modeled Arabic root-and-pattern morphology \cite{Beesley:1989:two-level,Kiraz:1994:multi-tape,Beesley:1998:arabic,Habash:2006:magead,Smrvz:2007:elixirfm}. A parallel line of work adopted lexicon- and compatibility-table-based resources, most notably the Buckwalter Arabic Morphological Analyzer and its successors \cite{Buckwalter:2002:buckwalter,Maamouri:2010:ldc}. Aragen/ALMORGEANA further extended Buckwalter-style lexical resources toward generation from lexeme-and-feature representations \cite{Habash:2005:morphological}. More recently, \citet{khairallah-etal-2024-camel} introduced  CamelMorph, a comprehensive open-source morphological analyzer and generator for MSA that builds on this tradition, reporting over 100K lemmas with rich morphological features and broad paradigm coverage. In this work, we use the  CamelMorph lexicon and database to construct our controlled benchmark set, and we use its generation engine, exposed through CAMeL Tools \cite{obeid-etal-2020-camel}, to generate and validate reference inflected forms.

\paragraph{LLM Evaluation for Morphological Inflection}
Recently, benchmarking LLMs for morphological generation, particularly inflection, has gained traction as researchers increasingly probe LLMs for linguistic knowledge. Early work in this area used variants of the Wug test \cite{Berko-1958-wug} to evaluate morphological productivity and generalization in LLMs across typologically diverse languages \cite{weissweiler-etal-2023-counting,anh-etal-2024-morphology}. \citet{ismayilzada-etal-2025-evaluating} extended this line by evaluating morphological compositional generalization through both generative and discriminative tasks. However, these studies do not focus on Arabic or on controlled generation from explicit Arabic morphosyntactic feature specifications, which is the focus of our work.

Very few studies have focused on Arabic morphological generation in LLMs. IMPACT \cite{saeed2025impactinflectionalmorphologyprobes} introduced a multilingual evaluation framework for inflectional morphology across five morphologically rich languages, including Arabic, but its Arabic component targets a narrower set of agreement and inflectional phenomena. The work by \citet{alakeel2026morphemesbordersevaluatingrootpattern} focuses solely on Arabic, specifically MSA, evaluating how LLM tokenizers align with Arabic morphological structure and how LLMs perform on productive root--pattern generation. It constructs a controlled test set with real and nonce roots and finds that tokenizer--morpheme alignment is neither necessary nor sufficient for successful morphological generation. In contrast, our work targets controlled Arabic morphological generation over an explicitly structured feature space, focusing on inflected-form realization and systematic sampling rather than root--pattern productivity alone, gender-focused generation, or broad multilingual probing. Furthermore, our benchmark dataset is substantially larger and more comprehensive.

\section{Linguistic Background \& Terminology}

Arabic morphology is characterized by both \textbf{richness} and \textbf{complexity}. Its richness stems from large inflectional paradigms and productive cliticization involving conjunctions, prepositions, articles, and pronouns, yielding many possible surface forms for a single lemma (Table~\ref{tab:lemma-counts}). Its complexity arises from interactions between templatic ({root} and {pattern}) and concatenative (affix or clitic) morphemes, often accompanied by orthographic and morphophonological alternations.

\paragraph{Relevant Terminology}

We briefly summarize the main terminology used throughout the paper, following CamelMorph \cite{khairallah-etal-2024-camel}. A \textbf{lemma} represents an abstraction over all inflectional forms of a lexical item \cite{habash-etal-2022-morphotactic}. A \textbf{root} is an abstract consonantal sequence encoding core lexical meaning, while a \textbf{pattern} specifies the vocalic and templatic structure used to derive grammatical forms. A \textbf{stem} is the form produced by combining roots and patterns before affixation. Morphological representation is further organized through functional features such as gender, number, case, and state, as well as \textbf{part-of-speech (POS)}, which specifies the grammatical category, and the \textbf{gloss}, which provides an English semantic description. Finally, the framework distinguishes between \textbf{affixes}, which realize core morphosyntactic features within the \textbf{baseword}, and \textbf{clitics}, which include conjunctions, prepositions, definite article, and object and possessive pronouns.

\paragraph{Example} Consider the noun lemma \<لَجْنَة>~~\textit{laj.na{\TAMARBUTA}} `committee' which illustrates multiple types of interactions. It is derived from the sound root \textit{l.j.n} using the singular pattern \textit{1a2.3a{\TAMARBUTA}}. Its plural is the broken plural \<لِجَان>~~\textit{lijaAn}, which preserves the root while changing the pattern to \textit{1i2aA3}. Clitic attachment further increases surface variation. For example, attaching the enclitic pronoun \<هَا> \textit{haA} `her' produces \<لَجْنَتُهَا>~~\textit{laj.n+at+u+haA} `her committee [nominative]', where \<ة>~\textit{a{\TAMARBUTA}} surfaces as \<تَ>~\textit{at}. 
%
Single proclitic attachment yields forms such as \<اللِّجَانِ> \textit{Al+l{\SHADDA}ijaAni} `the committees [genitive]' and \<لِلِجَانٍ> \textit{li+lijaAn{\KASRATAN}} `for committees [genitive]'. Combining both triggers orthographic assimilation, producing \<لِلِّجَانِ> \textit{li+l{\SHADDA}jaAni} `for the committees [genitive]'.
%

\section{{\yallamorph} Benchmark Design}


    

   

\begin{table*}[th]
\centering
\small
\setlength{\tabcolsep}{5pt}
\begin{tabularx}{\textwidth}{llll}
\toprule
\textbf{Category} & \textbf{Type} & \textbf{Explanation} & \textbf{POS Group} \\
\midrule

\multirow{3}{*}{\makecell[l]{\textbf{Lemma}\\\textbf{Frequency}}}
& High & High-frequency lemma & All \\
& Medium & Medium-frequency lemma & All \\
& Low & Low-frequency lemma & All \\

\midrule

\multirow{7}{*}{\makecell[l]{\textbf{Root}\\\textbf{Class}}}
& Sound & Root with no weak letters, hamza, or gemination & All \\
& Geminated & Root with identical second and third radicals & All \\
& Hamzated & Root containing a hamza consonant & All \\
& Weak Initial & Root whose first radical is weak (\textit{w} or \textit{y}) & All \\
& Hollow & Root whose second radical is weak & All \\
& Defective & Root whose third radical is weak & All \\
& NTWS & Non-templatic word stem & Nouns, Adjectives \\

\midrule

\multirow{3}{*}{\makecell[l]{\textbf{Paradigm}\\\textbf{Completeness}}}
& Full & Fully inflecting paradigm & All \\
& Masculine & Masculine only nominal paradigm & Nouns \\
& Feminine & Feminine only nominal paradigm & Nouns \\

\midrule

\multirow{7}{*}{\makecell[l]{\textbf{Stem}\\\textbf{Complexity}}}
& Regular & No major orthographic or morphological alternations & All \\
& Defective Family & Stem allomorphs exhibit defective-family alternations & All \\
& Hamza Family & Stem allomorphs exhibit hamza-related alternations & All \\
& \#N Family & Stem ends with letter \<ن> \textit{n}. & Verbs \\
& \#T Family & Stem ends with letter \<ت> \textit{t}. & Perfective Verbs \\
& L\# Family & Stem begins with letter \<ل> \textit{l}. & Nouns, Adjectives \\
& Diptote Family & At least one stem allomorph is a diptote \<ممنوع من الصرف>. & Nouns, Adjectives \\

\bottomrule
\end{tabularx}
\caption{Definitions of sampling categories used for lemma selection.}
\label{tab:sampling-category-definitions}
\end{table*}

In this section, we present the design details of the {\yallamorph} benchmark.

\subsection{Design Principles}

The design of {\yallamorph} is centered on evaluating Arabic morphological generation at scale while maintaining linguistic balance and interpretability. The benchmark includes over 600K sampled forms covering the major open POS classes of Arabic and capturing both inflectional and cliticization phenomena. To ensure broad linguistic coverage, the dataset is balanced across a wide range of morphophonological dimensions, including root classes, stem complexity, and paradigm completeness. In addition, {\yallamorph} incorporates frequency-balanced sampling over lemmas and morphological categories, enabling controlled evaluation of model generalization across both frequent and long-tail forms.

\subsection{Data Source}

We construct our benchmark from the lemma inventory and morphological analyses provided by  CamelMorph MSA~\cite{khairallah-etal-2024-camel}. Specifically, we use the lemma entries, part-of-speech labels, glosses, and morphological features to support sampling, categorization, and paradigm generation. Access to these analyses is handled through CAMeL Tools \cite{obeid-etal-2020-camel}.

\subsection{POS Groups \& Morphological Features}

{\yallamorph} focuses on the three major open POS classes in Arabic: verbs, nouns, and adjectives. Within the verbal domain, we distinguish five core configurations: active perfective, passive perfective, active imperfective, passive imperfective, and command. The various POS groups  differ in inflectional behavior, stem complexity, and compatibility with clitic attachment.
To capture both core morphology and cliticization attachment phenomena, the benchmark is organized into two complementary subsets: a \textit{baseword} subset focusing on inflectional morphology, and a \textit{cliticization} subset focusing on interactions between stems and attached clitics.

\subsection{Sampling Motivation}


Arabic morphology exhibits substantial lexical and inflectional richness. Combining the large lemma inventory of CamelMorph MSA with feature bundles and clitic configurations yields an impractically large number of possible test instances (Table~\ref{tab:lemma-counts}). We therefore construct the benchmark through guided sampling rather than exhaustive enumeration. The following subsections describe the linguistic and distributional dimensions underlying this sampling process. Our benchmark comprises over 600K carefully sampled entries.


\begin{table}[t]
\centering
\footnotesize
\begin{tabular}{lrrr}
\toprule
&& \multicolumn{2}{c}{\textbf{Hypothetical Maximum}}\\
\textbf{POS} 
& \textbf{Lemmas} & \textbf{- Clitics} & \textbf{+ Clitics} \\

\midrule
Verbs      &  9,333 & 2,183,454 &   754,417,944 \\
Nouns      & 19,844 & 1,071,576 & 2,730,375,648 \\
Adjectives &  6,924 &   373,896 &   952,687,008 \\
\midrule
\textbf{Total} & \textbf{36,101} & \textbf{3,628,926} & \textbf{4,437,480,600} \\
\bottomrule
\end{tabular}
\caption{Lemma and maximum instance counts by lexical class, with and without clitics.}
\label{tab:lemma-counts}
\vspace{-12pt}
\end{table}

\subsection{Sampling Dimensions}

We sample lemmas along linguistically motivated dimensions capturing root class, lemma frequency, paradigm completeness, and stem complexity. Each eligible lemma is assigned categorical labels for the relevant dimensions, and combinations of these labels define the sampling strata.
Table~\ref{tab:sampling-category-definitions} summarizes all category dimensions and their possible values, while Table~\ref{tab:representative-examples} presents representative noun and active perfective verb (PVA) examples illustrating their combinations.
 
\begin{table*}[t]
\centering
\small
\setlength{\tabcolsep}{4pt}
\renewcommand{\arraystretch}{0.0}
\setlength{\extrarowheight}{0pt}
\begin{tabularx}{\textwidth}{lllll r l l}
\toprule
\textbf{Group} & \textbf{Frequency} & \textbf{Root Class} &
\textbf{Paradigm} & \textbf{Stem Complexity} &
\multicolumn{2}{l}{\textbf{Example Lemma}} & \textbf{Gloss} \\
\midrule
Verb & High & Sound & Full & Regular & \<رَجَع> & \textit{raja{\AYN}} & return \\
Verb & High & Hamzated & Full & \#Hamza Family & \<اِبْتَدَأ> & \textit{Aib.tada{\AHAMZAUP}} & begin \\

Verb & Medium & Geminated & Full & \#N Family & \<اِمْتَنّ> & \textit{Aim.tan{\SHADDA}} & be grateful \\
Verb & Medium & Hollow & Full & Regular & \<نَاف> & \textit{naAf} & exceed \\

Verb & Low & Sound & Full & Regular & \<تَبَلْمَر> & \textit{tabal.mar} & be polymerized \\
Verb & Low & Defective & Full & Regular & \<رَخُو> & \textit{raxuw} & be loose \\

\midrule
Noun & High & Sound & Full & \#Hamza Family & \<شَرِيك> & \textit{{\SHIN}ariyk} & partner \\
Noun & High & NTWS & Masculine & L\# Family & \<لِيتْر> & \textit{liyt.r} & liter \\

Noun & Medium & Hollow & Feminine & \#Defective Family & \<صِينِيَّة> & \textit{Siyniy{\SHADDA}a{\TAMARBUTA}} & porcelain \\
Noun & Medium & Weak Initial & Masculine & \#Diptote Family & \<مَوْشُور> & \textit{maw.{\SHIN}uwr} & prism \\

Noun & Low & Sound & Feminine & L\# Family & \<لَااِنْقِسَامِيَّة> & \textit{laAAin.qisaAmiy{\SHADDA}a{\TAMARBUTA}} & indivisibility \\
Noun & Low & Geminated & Full & Regular & \<كَظِيظ> & \textit{ka{\ZA}iy{\ZA}} & overfilled \\

\bottomrule
\end{tabularx}
\caption{Representative verb and noun lemmas illustrating sampling categories across lemma frequency, root class, paradigm completeness, and stem complexity. Verb categories are from the perfective active subset.}
\label{tab:representative-examples}
\vspace{-6pt}
\end{table*}

%

\paragraph{Lemma Frequency}
Lemma frequency provides a corpus-based estimate of how often
each lemma is attested in a reference corpus. We estimate
frequency using BAREC-10M \citep{elmadani-etal-2026-large},
a large, balanced, multi-domain corpus of Arabic that is
automatically morphologically tagged and lemmatized.

Our lemma-frequency analysis was conducted in parallel with the
BAREC-10M annotation effort. 
The final BAREC-10M annotations and their additional
lemma-cluster resolution procedure were not used because they
were not yet available at the time of our analysis.
We therefore independently
analyzed the raw corpus using the morphological disambiguation
component of CAMeL Tools.\footnote{We used CAMeL Tools
v1.5.5 \citep{obeid-etal-2020-camel}, with BERT-Disambig
\citep{inoue-etal-2022-morphosyntactic} and the CamelMorph MSA v1.1
database \citep{khairallah-etal-2024-camel}.}
For each token, we retained the top-ranked morphological analysis.

Lemmas for which our analysis recovered no occurrences were
assigned a count of zero. For sampling, we group lemmas into
three frequency bands: low, medium, and high. The low band
contains lemmas with frequency 0; the medium band contains
lemmas with frequencies from 1 to 20; and the high band contains
lemmas with frequencies above 20.

\paragraph{Root Class}
Root class captures the phonological type of a lemma's root. We use seven root-class labels. Each lemma is assigned to exactly one root-class label. For lemmas with identifiable roots, overlapping classifications are resolved using the following priority hierarchy: \textit{geminated} $>$ \textit{hamzated} $>$ \textit{weak-initial} $>$ \textit{hollow} $>$ \textit{defective} $>$ \textit{sound}. Lemmas with no identifiable root are treated as non-templatic word stems (NTWS) and assigned to the \textit{NTWS} class. This procedure yields a single, consistent root-class label for each lemma and supports balanced sampling across major root types.



\paragraph{Paradigm Stem Complexity}
Paradigm stem complexity captures broad stem-related patterns that affect how forms are realized across a lemma's paradigm. 
We derive this dimension from stem-related information provided in  CamelMorph MSA.
We assign lemmas to one of seven stem-complexity families and some lemmas may satisfy {\textit{more than one}} stem-complexity condition. Overlapping classifications are resolved using the following priority hierarchy: \textit{\#N Family} $>$ \textit{\#T Family} $>$ \textit{L\# Family} $>$ \textit{Defective Family} $>$ \textit{Hamza Family} $>$ \textit{Diptote Family} $>$ \textit{Regular}. This procedure yields a single, consistent stem-complexity label for each lemma and supports non-overlapping sampling across stem-complexity families. For example, the lemma \<شَرِيك> \textit{{\SHIN}ariyk} `partner' has a broken plural \<شُرَكَاء> \textit{{\SHIN}urakaA'} which ends in a Hamza (glottal stop). This places the whole lemma in the Hamza Family group. Table~\ref{tab:sampling-category-definitions} defines the various stem-complexity families. Some of these apply to all POS, such as Hamza Family, while others are very specific, e.g., \#T~Family only applies to perfective verbs and captures the reduced spelling of t-initial suffixes with t-final verbs due to orthographic gemination (Shadda): \<فُتُّ> \textit{fut{\SHADDA}u} (\textit{fut+tu}) `I entered'.

 \paragraph{Paradigm Completeness} 
Paradigm completeness captures the extent to which a lemma realizes the expected range of gender--number combinations in its paradigm. 
We derive this dimension from stem-related information provided in  CamelMorph MSA.
Lemmas are assigned to one of three labels: \textit{masculine-only}, \textit{feminine-only} and  \textit{full}.
%
%
Verbal and adjectival paradigms are always \textit{full} in Arabic, whereas nouns can be of any of the three types.
The inclusion of \textit{masculine-only} and \textit{feminine-only} paradigms leads to the possibility of Null forms, i.e., feature combinations that cannot be accommodated such as the feminine of \<مَكْتَب> \textit{mak.tab} `office' is a masculine-only lemma. Simply adding the feminine ending to produce \<مَكْتَبَة> \textit{mak.taba{\TAMARBUTA}} `library' yields an incorrect result. Around 13\% of all the entries in {\yallamorph} correctly have a \textit{null} gold reference.

\subsection{Lemma Sampling Process}

After assigning sampling dimension labels to all lemmas, we group them into strata defined by joint combinations of the sampling dimensions across the benchmark POS groups. Since these strata vary greatly in size, we adopt a fixed per-stratum strategy, sampling up to 10 lemmas from each non-empty stratum while including all lemmas for smaller strata for the baseword subset of the benchmark. This preserves broad linguistic coverage while limiting over-representation of highly productive categories.
The cliticization subset is constructed separately due to the large expansion introduced by proclitic and enclitic combinations. Instead of exhaustively enumerating all cliticized forms, we derive a reduced sample from the baseword subset while preserving coverage across lemma frequency and stem complexity, and we only include full paradigm completeness cases. For each relevant category, we sample up to 5 lemmas, yielding a manageable yet linguistically diverse clitic-focused evaluation set.

\subsection{Arabic Morphological Generation Entries}

We formulate the task as slot-level Arabic morphological generation. Given a lemma, POS, gloss, and a target feature bundle (henceforth, \textit{morphological specification}), the model generates the corresponding inflected surface form. Outputs may consist of a single valid form, multiple valid realizations, or an indication that the requested configuration is invalid (\textit{null} reference).

Multiple-reference cases account for 3.2\% of the benchmark entries. In these cases, the gold reference set contains more than one valid realization for the same morphological specification.

%
Morphological specifications are POS-dependent. Verbal forms are specified using aspect, person, gender, number, voice, and mood, while nouns and adjectives use gender, number, case, and state. Clitic configurations are represented separately using positional proclitic (\textit{prc0--3}) and enclitic (\textit{enc0--1}) slots, with POS-specific constraints. The full list of features is provided in Appendix~\ref{app:morph-feats}. The selected clitic inventory and  compatibility restrictions are provided in
Appendix~\ref{app:clitic-features}.


\begin{table}[t]
\centering
\small
\setlength{\tabcolsep}{4pt}
\begin{tabular}{lrrrr}
\toprule
& \multicolumn{2}{c}{\textbf{Lemmas}} & \multicolumn{2}{c}{\textbf{Inflected Forms}} \\
\cmidrule(lr){2-3} \cmidrule(lr){4-5}
\textbf{POS Group} & \textbf{Baseword} & \textbf{Clitics} & \textbf{Baseword} & \textbf{Clitics} \\
\midrule
Active PV   & 513   & 75 & 9,234   & 45,738  \\
Passive PV  & 498   & 70 & 8,964   & 8,820   \\
CV          & 433   & 60 & 7,794  & 6,480   \\
Active IV   & 435   & 60 & 39,150  & 181,800 \\
Passive IV  & 435   & 60 & 39,150  & 37,800  \\
Nouns       & 1,887 & 75 & 101,898 & 67,950  \\
Adj         & 594   & 75 & 32,076  & 76,950  \\
\midrule
Subtotal    & 4,795 & 475 & 238,266 & 425,538 \\
\midrule
Total       & \multicolumn{2}{c}{4,795} & \multicolumn{2}{c}{663,804}  \\
\bottomrule
\end{tabular}
\caption{Benchmark subset sizes.}
\label{tab:benchmark-sizes}
\vspace{-12pt}
\end{table}

\subsection{Benchmark Statistics}

\begin{table}[t]
\centering
\small
\begin{tabular}{lrr}
\toprule
\textbf{Frequency} & \textbf{Instance Count} & \textbf{Percentage} \\
\midrule
10M+        & 43      & 0.0 \\
1M--10M     & 721     & 0.1 \\
100K--1M    & 5,040   & 0.8 \\
10K--100K   & 14,068  & 2.1 \\
1K--10K     & 25,699  & 3.9 \\
101--1K     & 36,022  & 5.4 \\
11--100     & 43,746 & 6.6 \\
1--10       & 55,908  & 8.4 \\
0           & 397,179 & 59.8 \\
Null        & 85,378  & 12.9 \\
\midrule
Total       & 663,804 & 100 \\
\bottomrule
\end{tabular}
\caption{Frequency distribution of entries.}
\label{tab:frequency_distribution}
\vspace{-12pt}
\end{table}

Table~\ref{tab:benchmark-sizes} summarizes benchmark sizes across POS groups and experimental subsets. Although the cliticization subsets contain fewer lemmas, they are substantially larger in total size due to the multiplicative effect of generating all inflected baseword forms and their cliticized variants.
Table~\ref{tab:frequency_distribution} shows the frequency distribution of undiacritized target forms based on the CAMeLBERT Frequency List~\cite{Khalifa:2021:Camel_Frequency}, highlighting the strong long-tail nature of the benchmark, with
 59.8\% of unseen forms, while an additional 12.9\% correspond to null reference entries.

\section{Evaluation}

\subsection{Metrics}
Our main metric is \textbf{Any Match Accuracy}, computed at the
instance level by comparing each prediction against its corresponding
gold form or set of gold forms. For multi-form outputs, AMA counts a
prediction as correct if at least one of the generated forms matches a
gold form. Target configurations with a Null reference, representing
morphologically invalid lemma--feature combinations, are counted as
correct only when the model explicitly predicts a Null output.

We additionally report micro-averaged \textbf{Precision},
\textbf{Recall}, and \textbf{F1} to capture partial correctness in
multi-form outputs. We evaluate predictions under three orthographic
settings: \textit{diacritized}, \textit{undiacritized}, and
\textit{Alif/Ya/Hamza/Ta-Marbuta normalized}. We treat the diacritized
setting as primary because it provides the strictest evaluation of
Arabic morphological generation. Table~\ref{tab:overall-results}
reports AMA and F1, while the corresponding Precision and Recall
results are provided in Appendix~\ref{app:precision_recall}.

\begin{table*}[t]
\centering
\small
\setlength{\tabcolsep}{4pt}

\begin{tabular}{l*{12}{r}}
\toprule
& \multicolumn{6}{c}{\textbf{10-Shot}}
& \multicolumn{6}{c}{\textbf{Zero-Shot}} \\
\cmidrule(lr){2-7}
\cmidrule(lr){8-13}

& \multicolumn{3}{c}{\textbf{Any Match Accuracy}}
& \multicolumn{3}{c}{\textbf{F1 Score}}
& \multicolumn{3}{c}{\textbf{Any Match Accuracy}}
& \multicolumn{3}{c}{\textbf{F1 Score}} \\
\cmidrule(lr){2-4}
\cmidrule(lr){5-7}
\cmidrule(lr){8-10}
\cmidrule(lr){11-13}

\textbf{Model}
& \multicolumn{1}{c}{\textbf{Diac}}
& \multicolumn{1}{c}{\textbf{Undiac}}
& \multicolumn{1}{c}{\textbf{Norm}}
& \multicolumn{1}{c}{\textbf{Diac}}
& \multicolumn{1}{c}{\textbf{Undiac}}
& \multicolumn{1}{c}{\textbf{Norm}}
& \multicolumn{1}{c}{\textbf{Diac}}
& \multicolumn{1}{c}{\textbf{Undiac}}
& \multicolumn{1}{c}{\textbf{Norm}}
& \multicolumn{1}{c}{\textbf{Diac}}
& \multicolumn{1}{c}{\textbf{Undiac}}
& \multicolumn{1}{c}{\textbf{Norm}} \\
\midrule

\textbf{GPT}$_{en}$
& \textbf{51.9} & \textbf{67.0} & \textbf{67.7}
& \textbf{54.4} & \textbf{70.5} & \textbf{71.3}
& \textbf{46.2} & \textbf{61.6} & \textbf{62.7}
& \textbf{49.2} & \textbf{66.0} & \textbf{67.3} \\

\textbf{Gemini}$_{en}$
& 49.7 & 60.7 & 61.4
& 52.4 & 64.2 & 64.9
& 39.1 & 53.7 & 54.3
& 41.5 & 57.0 & 57.7 \\

\textbf{Fanar}$_{en}$
& 21.5 & 36.5 & 37.2
& 22.5 & 38.3 & 39.1
& 5.1 & 19.4 & 19.9
& 5.3 & 20.3 & 20.9 \\

\textbf{Fanar}$_{ar}$
& 15.8 & 26.8 & 27.2
& 16.6 & 28.1 & 28.5
& 2.7 & 15.5 & 15.8
& 2.8 & 16.1 & 16.5 \\

\textbf{Jais-70B}$_{en}$
& 9.1 & 20.9 & 21.2
& 9.5 & 21.7 & 22.1
& 8.2 & 15.6 & 15.9
& 8.8 & 16.5 & 16.9 \\

\textbf{Jais-70B}$_{ar}$
& 8.2 & 17.5 & 17.8
& 8.6 & 18.3 & 18.6
& 3.8 & 10.7 & 11.3
& 4.1 & 11.4 & 12.1 \\

\textbf{ALLaM}$_{en}$
& 4.7 & 10.7 & 11.0
& 4.7 & 10.9 & 11.2
& 4.4 & 8.6 & 8.8
& 3.5 & 7.2 & 7.3 \\

\textbf{ALLaM}$_{ar}$
& 4.5 & 10.0 & 10.3
& 4.1 & 9.3 & 9.5
& 3.1 & 9.0 & 9.2
& 2.8 & 8.4 & 8.6 \\

\textbf{Qwen}$_{en}$
& 4.5 & 13.0 & 13.6
& 4.9 & 14.4 & 15.0
& 2.0 & 8.7 & 9.1
& 2.1 & 9.2 & 9.7 \\

\textbf{Jais-8B}$_{en}$
& 3.4 & 10.1 & 10.2
& 3.3 & 9.9 & 10.0
& 2.7 & 6.4 & 6.7
& 2.3 & 5.7 & 6.0 \\

\textbf{Jais-8B}$_{ar}$
& 2.3 & 8.2 & 8.5
& 2.1 & 7.6 & 7.9
& 2.6 & 7.7 & 7.9
& 2.1 & 6.4 & 6.6 \\

\bottomrule
\end{tabular}

\caption{Overall benchmark results under the 10-shot and zero-shot
prompting settings. Subscripts \textit{ar} and \textit{en} indicate
the prompt language.}
\label{tab:overall-results}
\vspace{-12pt}
\end{table*}

\subsection{Compared Models}


We evaluate seven instruction-tuned LLMs spanning commercial multilingual, open-weight multilingual, and Arabic-focused model families: \textbf{GPT-5.4}, \textbf{Gemini 3.1 Flash-Lite}, \textbf{Qwen2.5-7B-Instruct}, \textbf{Fanar-2-27B-Instruct}, \textbf{Jais-2-8B-Chat}, \textbf{Jais-2-70B-Chat} and \textbf{ALLaM-7B-Instruct-preview}\footnote{Abbreviations used throughout: GPT-5.4 (GPT), Gemini 3.1
Flash-Lite (Gemini), Qwen2.5-7B-Instruct (Qwen),
Fanar-2-27B-Instruct (Fanar), Jais-2-70B-Chat (Jais-70B),
Jais-2-8B-Chat (Jais-8B), and ALLaM-7B-Instruct-preview
(ALLaM).}. This selection provides broad coverage of widely used general-purpose systems and models developed specifically for Arabic.

\paragraph{Prompt Design}

We use POS-specific prompts for verbs, nouns, and adjectives, with separate variants for clitic-aware experiments. Inputs include the lemma, POS, gloss, and the relevant morphological feature bundle.
Models are instructed to return outputs in a fixed JSON schema. We evaluate both \textbf{zero-shot} and \textbf{few-shot} prompting settings, with the few-shot setting including 10 in-context examples. Prompts remain lightweight, with only minimal clarifications for features that showed consistent ambiguity in pilot experiments, such as nominal state distinctions, emphatic verbal moods, and clitic ordering.

 We used English and Arabic versions of the prompts (Appendix~\ref{app:prompts}). Arabic and English prompts are used for \textbf{ALLaM}, \textbf{Fanar},
\textbf{Jais-70B} and \textbf{Jais-8B}; all other models use English prompts only.

\paragraph{Output Processing}
Model outputs are parsed using a unified post-processing pipeline that extracts candidate Arabic strings, handles minor formatting deviations from the requested JSON schema, and removes duplicate forms.

\subsection{Results}

We first compare overall model performance, then analyze the best-performing model across linguistic and distributional dimensions.
\paragraph{Overall Performance}
Table~\ref{tab:overall-results} shows that GPT substantially outperforms all other models across both Any Match Accuracy and F1 metrics in all evaluation settings, followed by Gemini. In contrast, the Arabic-oriented open models perform considerably worse overall, suggesting that Arabic specialization alone does not guarantee accurate morphological generation. English prompts generally perform better, especially in the 10-shot setting. However, Arabic prompts perform better for ALLaM and Jais-8B on several zero-shot undiacritized and normalized metrics.

\begin{table}[t]
\centering
\small
\setlength{\tabcolsep}{1.2pt}
\renewcommand{\arraystretch}{0.95}

\begin{tabular}{llcccccc}
\toprule
& & \multicolumn{3}{c}{\textbf{Base}}
& \multicolumn{3}{c}{\textbf{Clitics}} \\
\cmidrule(lr){3-5} \cmidrule(lr){6-8}

\textbf{Category} & \textbf{Group}
& \textbf{Diac} & \textbf{Und.} & \textbf{Norm}
& \textbf{Diac} & \textbf{Und.} & \textbf{Norm} \\
\midrule

\textbf{All} & All
& 57.4 & 67.9 & 68.3
& 48.8 & 66.6 & 67.4 \\
\midrule

\multirow{7}{*}{\makecell[c]{\textbf{POS}\\\textbf{Group}}}
& Active PV
& 79.1 & \textbf{88.9} & \textbf{89.2}
& 58.6 & 73.1 & 73.7 \\
& Passive PV
& 56.7 & 82.3 & 83.3
& 52.4 & \textbf{86.5} & \textbf{88.9} \\
& CV
& 45.7 & 74.6 & 78.8
& 21.7 & 46.9 & 55.4 \\
& Active IV
& 60.4 & 78.2 & 78.8
& 37.4 & 57.7 & 58.5 \\
& Passive IV
& 60.5 & 78.2 & 78.8
& 56.5 & 77.3 & 78.6 \\
& Nouns
& 46.2 & 49.7 & 49.8
& \textbf{69.8} & 80.4 & 80.9 \\
& Adj
& \textbf{82.3} & 88.7 & 89.0
& 49.5 & 65.5 & 65.8 \\
\midrule

\multirow{3}{*}{%
  \makecell[c]{%
    \textbf{LF}\\[-1pt]
    {\tiny Lemma}\\{\tiny Frequency}%
  }%
}
& High
& \textbf{59.2} & \textbf{69.3} & \textbf{69.8}
& \textbf{49.6} & \textbf{67.8} & \textbf{68.6} \\
& Medium
& 57.4 & 68.2 & 68.7
& 48.6 & 65.6 & 66.5 \\
& Low
& 55.1 & 65.7 & 66.1
& 48.1 & 66.2 & 67.0 \\
\midrule

\multirow{7}{*}{%
  \makecell[c]{%
    \textbf{RC}\\[-1pt]
    {\tiny Root Class}%
  }%
}
& Sound
& \textbf{62.2} & 70.5 & 70.6
& 55.8 & 72.0 & 72.2 \\
& Geminated
& 57.0 & 68.2 & 68.4
& 46.8 & 66.7 & 66.8 \\
& Hamzated
& 56.9 & 67.1 & 68.7
& 43.3 & 58.7 & 61.0 \\
& Weak Initial
& 54.5 & 69.8 & 70.2
& 50.7 & 70.4 & 70.7 \\
& Hollow
& 59.7 & \textbf{70.6} & \textbf{70.8}
& 45.2 & 67.0 & 67.5 \\
& Defective
& 56.2 & 65.8 & 66.3
& 53.2 & 69.9 & 70.6 \\
& NTWS
& 49.1 & 55.6 & 55.6
& \textbf{69.2} & \textbf{82.5} & \textbf{82.5} \\
\midrule

\multirow{3}{*}{%
  \makecell[c]{%
    \textbf{PC}\\[-1pt]
    {\tiny Paradigm}\\{\tiny Completeness}}%
  }%

& Full
& \textbf{67.8} & \textbf{81.9} & \textbf{82.5}
& 48.8 & 66.6 & 67.4 \\
& Masculine
& 33.6 & 37.1 & 37.3
& -- & -- & -- \\
& Feminine
& 39.5 & 42.2 & 42.2
& -- & -- & -- \\
\midrule

\multirow{7}{*}{%
  \makecell[c]{%
    \textbf{SC}\\[-1pt]
    {\tiny Stem}\\\tiny{Complexity}%
  }%
}
& Regular
& \textbf{62.3} & 72.4 & 72.7
& 55.5 & 73.6 & 73.8 \\
& Defective
& 53.7 & 67.5 & 68.2
& 46.0 & 66.8 & 67.5 \\
& Hamza
& 55.1 & 61.3 & 62.7
& 44.3 & 57.0 & 59.4 \\
& \#T
& 61.7 & \textbf{79.2} & 79.1
& 39.8 & 68.2 & 68.2 \\
& \#N
& 60.0 & 79.1 & \textbf{79.6}
& 42.5 & 65.7 & 65.9 \\
& L\#
& 55.1 & 59.3 & 59.4
& \textbf{62.1} & 72.3 & 72.4 \\
& Diptote
& 49.0 & 51.2 & 51.2
& 61.7 & \textbf{75.5} & \textbf{75.5} \\
\bottomrule
\end{tabular}

\caption{Detailed GPT results across benchmark dimensions under the 10-shot prompting setting.}
\label{tab:gpt_detailed_fewshot_results}
\vspace{-12pt}
\end{table}

\paragraph{Detailed Results}
We consider the detailed results for GPT, the best-performing model under the English 10-shot setting.
Table~\ref{tab:gpt_detailed_fewshot_results} shows substantial differences between the baseword and cliticization settings, particularly under diacritized evaluation, whereas the differences are smaller in the undiacritized and normalized settings. Across POS groups, adjectives and active perfective verbs perform best in the baseword setting, while nouns achieve the highest cliticized diacritized accuracy and passive perfective verbs lead under undiacritized and normalized evaluation. Lemma frequency has a relatively modest association with performance, whereas root-class differences are more pronounced under cliticization. Full paradigms outperform masculine-only and feminine-only paradigms, partly reflecting the difficulty of Null-reference configurations. Stem-Complexity rankings also vary across settings: \#T and \#N perform strongly on undiacritized basewords but decline under cliticization, while L\# and Diptote families perform comparatively well in the clitic subset.

\begin{table}[t]
\centering
\small
\setlength{\tabcolsep}{4pt}
\begin{tabular}{lrrrr}
\toprule
\textbf{Frequency} & \textbf{Examples} &
\textbf{Diac} & \textbf{Undiac} & \textbf{Norm} \\
\midrule
10M+       & 43      & 81.3 & 88.4 & 93.0 \\
1M--10M    & 721     & 79.5 & 90.2 & 90.6 \\
100K--1M   & 5,040   & 79.4 & 88.1 & 88.3 \\
10K--100K  & 14,068  & 77.2 & 88.0 & 88.2 \\
1K--10K    & 25,699  & 73.8 & 86.3 & 86.7 \\
101--1K    & 36,022  & 69.5 & 84.1 & 84.6 \\
11--100    & 43,746  & 68.1 & 83.4 & 84.1 \\
1--10      & 55,908  & 65.5 & 81.6 & 82.3 \\
0          & 397,179 & 52.3 & 71.1 & 72.0 \\
Null       & 85,378  & 12.5 & 12.5 & 12.5 \\
\midrule
$>0$       & 181,247 & 69.5 & 83.9 & 84.5 \\
All        & 663,804 & 51.9 & 67.0 & 67.7 \\
\bottomrule
\end{tabular}
\caption{GPT 10-shot Any Match Accuracy (AMA) across surface-form frequency bands derived from  CAMeLBERT Frequency.}
\label{tab:gpt_frequency_results}
\end{table}

\paragraph{Performance Across Surface Form Distributions}
Table~\ref{tab:gpt_frequency_results} shows a strong correlation between frequency and performance. GPT achieves high accuracy on frequent forms, with performance gradually decreasing toward rarer forms. Null-reference configurations are the most difficult category, with 12.5\% accuracy, and account for 12.9\% of the benchmark. In contrast, performance on attested forms (\textit{>0}) remains considerably higher, suggesting that Arabic morphological generation in LLMs is still strongly tied to exposure frequency and memorization, despite moderate generalization across low-frequency forms.

\begin{table}[t]
\centering
\small
\setlength{\tabcolsep}{2pt}
\begin{tabular}{ccrrrr}
\toprule
\makecell{\textbf{Feature-Freq}\\\textbf{Percentile}} & \textbf{Examples} & \textbf{\%}
& \textbf{Diac} & \textbf{Undiac} & \textbf{Norm} \\
\midrule
75--100\% & 13,236  & 2.0  & 65.2 & 80.1 & 80.2 \\
25--75\%  & 59,438  & 9.0  & 55.4 & 69.8 & 70.0 \\
0--25\%   & 356,131 & 53.7 & 55.2 & 67.9 & 68.5 \\
0          & 234,999 & 35.4 & 45.2 & 64.2 & 65.3 \\
\midrule
\textbf{All} & 663,804 & 100
& 51.9 & 67.0 & 67.7 \\
\bottomrule
\end{tabular}
\caption{GPT-5.4 10-shot Any Match Accuracy (AMA) by
feature-frequency percentile, computed from BAREC-10M
counts of POS-specific morphological specifications
excluding the lemma. Zero-count specifications form a
separate group; nonzero specifications are ranked by count
and divided into the shown percentile bands.}
\label{tab:gpt_morphspec_frequency}
\vspace{-12pt}
\end{table}

\paragraph{Performance Across Morphological-Specification Distributions}

Table~\ref{tab:gpt_morphspec_frequency} shows that performance strongly correlates with the corpus frequency of morphological specifications. GPT performs best on highly frequent feature configurations and degrades steadily toward rarer specifications. Completely unseen specifications are substantially harder, especially in the diacritized setting. Unlike lemma-frequency effects, these results suggest that models are highly sensitive not only to lexical exposure, but also to the distributional frequency of underlying morphosyntactic patterns.

Overall, the results show that controlled Arabic morphological generation remains challenging even for strong modern LLMs, particularly for cliticized, unseen, and morphologically rare configurations. The findings further suggest that current models rely heavily on distributional exposure and struggle to robustly generalize across the full Arabic morphological space.







\begin{table}[htbp]
    \centering
\small
\setlength{\tabcolsep}{3pt}
    \begin{tabular}{lcccccc}
        \toprule
        \textbf{Model}
        & \textbf{LPOS}
        & \textbf{PGN}
        & \textbf{CS}
        & \textbf{VAM}
        & \textbf{PRC}
        & \textbf{ENC} \\
        \midrule
        \textbf{GPT (En)}
        & \textbf{75.0} & \textbf{55.2} & \textbf{75.0} & \textbf{73.3} & \textbf{75.7} & \textbf{77.1} \\

        \textbf{Gemini (En)}
        & 68.0 & 52.8 & 71.0 & 67.6 & 65.7 & 71.6 \\

        \textbf{Fanar (En)}
        & 67.2 & 43.6 & 71.9 & 61.9 & 51.1 & 69.2 \\

        \textbf{Fanar (Ar)}
        & 70.2 & 41.5 & 73.3 & 60.8 & 41.7 & 66.7 \\

        \textbf{Jais-70B (En)}
        & 63.4 & 37.1 & 69.6 & 53.1 & 35.3 & 55.4 \\

        \textbf{Jais-70B (Ar)}
        & 59.1 & 40.2 & 66.9 & 55.6 & 30.6 & 48.8 \\

        \textbf{ALLaM (En)}
        & 69.0 & 37.0 & 69.1 & 45.2 & 28.0 & 49.4 \\

        \textbf{ALLaM (Ar)}
        & 64.2 & 36.8 & 69.9 & 60.4 & 32.6 & 51.8 \\

        \textbf{Qwen (En)}
        & 50.5 & 25.7 & 62.0 & 39.5 & 38.8 & 50.4 \\

        \textbf{Jais-8B (En)}
        & 68.1 & 31.8 & 72.9 & 47.7 & 33.8 & 52.2 \\

        \textbf{Jais-8B (Ar)}
        & 70.9 & 34.1 & 72.6 & 53.9 & 30.4 & 51.5 \\
        \bottomrule
    \end{tabular}
        \caption{Comparison of models across morphosyntactic features: lemma-part-of-speech (LPOS), person–gender–number (PGN), case–state (CS), voice–aspect–mood (VAM), proclitics (PRC), and enclitics (ENC).}
    \label{tab:feature_group_errors}
    \vspace{-12pt}
\end{table}

\subsection{Error Analysis}





\paragraph{Qualitative Analysis} We randomly sampled 100 incorrect GPT outputs from the 10-shot baseword setting and another 100 incorrect
outputs from the 10-shot cliticization setting for
manual analysis.
In the baseword setting, 41\% of errors
involved generating forms for invalid configurations (instead of \emph{null}), 29\%
were diacritization-only differences, 19\% involved
incorrect stems, incorrect affixes or both, 10\% were
missing valid outputs, and 1\% was a Tatweel artifact.

In the cliticization setting, 39\% were diacritization-only differences, 39\% 
involved
incorrect stems, incorrect affixes or both,
13\% involved generating forms for invalid configurations, 5\% were missing valid outputs, and 4\% involved an incorrect clitic sequence form or order.

These results suggest that clitics were rarely the direct source of error. Instead, cliticization mainly increased the difficulty of realizing the host word correctly.

\paragraph{Quantitative Analysis} We examine feature-group recovery for all models in the 10-shot setting. This analysis shows which types of morphological information are successfully preserved and where errors are concentrated. Each generated form is normalized using the same procedure as in the main evaluation and then analyzed using the same CAMeL Tools/CamelMorph MSA configuration used for benchmark construction. A group is recovered if at least one analysis matches the target values of all its features. We examine six groups: lemma-part-of-speech (\texttt{LPOS}); person, gender, and number (\texttt{PGN}); case and state (\texttt{CS}); voice, aspect, and mood (\texttt{VAM}); proclitics (\texttt{PRC}); and enclitics (\texttt{ENC}). Table~\ref{tab:feature_group_errors} reports the recovery rates across systems.

The feature-group results should be interpreted alongside the
output-cardinality distribution reported in Table~\ref{tab:output-cardinality}. This is particularly important for Jais-8B, which generates multiple forms for 27.2\% and 17.6\% of
instances under English and Arabic 10-shot prompting, respectively,
compared with only 3.2\% of the references. Because feature-group
recovery assigns credit when at least one generated candidate has an
analysis matching the target features, this over-generation increases
the probability of recovering the correct lemma and POS. It may
therefore partly explain the relatively high \texttt{LPOS} recovery
of Jais-8B, particularly its score of 70.9\% under Arabic prompting,
despite its weaker recovery of other feature groups. However, generating
additional candidates can also introduce incorrect forms, resulting in
lower precision and F1. Feature-group recovery should therefore be
interpreted as a measure of morphological coverage rather than exact
generation accuracy.

\paragraph{Problematic Output Behavior}
Inspection of the complete test set revealed three recurring failure
modes.
First, Jais-8B (En, 0-shot) returned
an \textbf{unrelated default} \<كتاب> \textit{ktAb} `book' in 30,085 instances (4.5\%), regardless of the requested analysis. Jais-70B did not exhibit this behavior.
%
%
Second, Jais-70B~(Ar) produced \textbf{romanized forms} in 3,473 outputs (0.52\%), e.g.
\emph{ashrafu} and
\texttt{\textgreater{}a\_wa\_la\_na\_athara}. Jais-70B (En) produced only five romanized
outputs.
%
%
Finally, Fanar (0-shot) occasionally generated Quranic material
\textbf{unrelated to the requested morphological form}, e.g., the isolated sequence
\<الم> \textit{alif-lam-mim}, which opens
Surat al-Baqarah, appeared 2,374 times 
(0.36\%) in Fanar (En) and 3,318 times (0.50\%) in Fanar (Ar).

\section{Conclusion and Future Work}


We presented {\yallamorph}, a large-scale benchmark for controlled Arabic morphological generation covering verbs, nouns, adjectives, cliticized forms, and invalid configurations. The benchmark combines broad linguistic coverage with controlled sampling across frequency, root class, paradigm completeness, and stem complexity.
Our evaluation shows that Arabic morphological generation remains challenging for current LLMs, especially in diacritized and cliticized settings. Performance drops substantially for zero-frequency forms and rare morph specifications, suggesting strong reliance on distributional exposure and memorization rather than robust morphological generalization.

Future work will extend {\yallamorph} to additional Arabic varieties and contextual generation settings, while also enabling finer-grained analysis of errors involving agreement, stem selection, diacritization, and cliticization.

\section*{Limitations}

This work focuses on Modern Standard Arabic as represented in  CamelMorph MSA, and therefore does not cover the full diversity of Arabic dialects or mixed-register usage. The benchmark is also constrained by the coverage, analyses, and generation decisions of the underlying morphological resource. Although we include a large number of forms, the benchmark is still based on guided sampling rather than exhaustive enumeration of the full morphological space.

Because our lemma-frequency analysis was conducted in parallel
with the BAREC-10M annotation effort, the final released
annotations and associated lemma-cluster resolution procedure
were not available at the time of our analysis. The reported lemma
frequencies therefore reflect our independent analysis pipeline
and may differ from frequencies derived from the final
BAREC-10M annotations.

Our evaluation uses prompt-based generation with selected LLMs and prompt variants. Results may vary with alternative prompting strategies, decoding settings, or model versions. Finally, while exact-match evaluation is appropriate for controlled generation, it may not capture all cases of acceptable orthographic variation or context-dependent preference among valid forms.

\section*{Ethics Statement}
This work introduces a benchmark for evaluating Arabic morphological generation. The dataset is derived from existing linguistic resources and does not contain private, personal, or user-generated sensitive information. The benchmark is intended to support more accurate and linguistically informed Arabic NLP systems.

Potential risks include over-reliance on Modern Standard Arabic as a representative form of Arabic and the use of benchmark results to make broad claims about Arabic language competence. Arabic is highly diverse across regions and registers, and performance on {\yallamorph} should not be interpreted as full coverage of Arabic linguistic ability. We encourage responsible use of the benchmark alongside evaluations for dialectal Arabic, downstream tasks, and human-centered applications.

We used AI writing assistance within the scope of ``Assistance
purely with the language of the paper'' described in the ACL Policy on Publication Ethics.

\section*{Acknowledgments}
This research was conducted using the High Performance Computing resources at New York University Abu Dhabi (NYUAD). We gratefully acknowledge the Center of Interdisciplinary Data Science and AI (CIDSAI) at NYUAD for its generous support. This work was supported in part by Google.org and the Google Cloud Research Credits program through the Gemini Academic Program.
We sincerely thank our colleagues at CAMeL Lab, Ossama Obeid, Khalid Elmadani, and Mostafa Saeed, for their helpful conversations and support.

\bibliography{custom,camel-bib-v3,anthology,EMNLP_ArabicNLP_2026}

\appendix
\section{Morphological Features}
\label{app:morph-feats}
\subsection{Baseword Morphological Features}

The morphological feature space is POS-dependent: target slots for verbs, nouns, and adjectives are specified using different feature sets, as summarized in Table~\ref{tab:morph_features}. Verb slots are defined by \textit{aspect}, \textit{person}, \textit{gender}, \textit{number}, \textit{voice}, and \textit{mood}.

Noun and adjective slots are defined by \textit{gender}, \textit{number}, \textit{case}, and \textit{state}. 

\subsection{Clitic Morphological Features}
\label{app:clitic-features}
In addition to the core baseword morphological features, the benchmark models \textit{clitic configuration} as a separate source of variation. We represent clitics using positional dimensions for proclitic slots (\textit{prc0}, \textit{prc1}, \textit{prc2}, \textit{prc3}) and enclitic slots (\textit{enc0}, \textit{enc1}). These dimensions are POS-sensitive: not all clitic positions are available for all parts of speech. For example, verbal forms do not use all nominal values of \textit{prc0}, and nouns do not use the verbal enclitic configuration associated with \textit{enc1}. 

The selected clitic features and values used in our experiments are summarized in Table~\ref{tab:clitics}.

The full clitic space is very large when combined with core morphological feature bundles. We therefore use a restricted subset of clitic configurations rather than enumerating all combinatorially possible configurations. This subset is designed to preserve linguistically informative variation while keeping the evaluation feasible in terms of cost and runtime.

The selected clitics balance representativeness and coverage. For proclitics, we choose representative values, such as \texttt{wa\_conj} and \texttt{li\_prep}, whose attachment behavior is shared by other proclitics in their respective positional classes. For enclitics, we select values that cover different person, gender, and number combinations, as well as the main orthographic changes associated with attachment. Thus, while the selected clitics do not exhaust the full inventory, they cover the major attachment behaviors.

In particular, we retain selected values for clitic dimensions across verbs, nouns, and adjectives, and impose additional restrictions on nominal and adjectival configurations through interactions with features such as \textit{state}.

This reduction substantially decreases the number of tested configurations. For example, in active imperfective verbs, the unrestricted clitic space would yield 20 configurations for intransitive verbs, 100 for transitive verbs, and 500 for ditransitive verbs. Under the restricted setup, these numbers are reduced to 8, 40, and 168, respectively. These reductions make the clitic-focused evaluation practically manageable while preserving sufficient variation to analyze clitic-sensitive generation.

\begin{table}[t]
\centering
\small

\begin{tabular}{p{0.2\linewidth} p{0.62\linewidth}}
\toprule
\textbf{Verb } & \textbf{} \\
\textbf{ Feature} & \textbf{Values} \\

\midrule
person & 1st, 2nd, 3rd \\
gender & masculine, feminine \\
number & singular, dual, plural \\
aspect & perfective, imperfective, command \\
voice  & active, passive \\
mood   & indicative, subjunctive, jussive, energetic (light), energetic (heavy) \\
\bottomrule
\end{tabular}

\vspace{0.8em}

\begin{tabular}{p{0.2\linewidth} p{0.62\linewidth}}
\toprule
\textbf{Noun/Adj  } & \textbf{ } \\
\textbf{  Feature} & \textbf{Values} \\
\midrule
gender & masculine, feminine \\
number & singular, dual, plural \\
case   & nominative, accusative, genitive \\
state  & construct, definite, indefinite \\
\bottomrule
\end{tabular}

\caption{Core baseword morphological features and their possible values for verbs, nouns, and adjectives.}
\label{tab:morph_features}
\end{table}

\begin{table}[!t]
\centering
\small

\begin{tabular}{p{0.2\linewidth} p{0.68\linewidth}}
\toprule
\textbf{Verb } & \textbf{} \\
\textbf{ Clitic} & \textbf{Values} \\
\midrule
prc1 & 0, la\_rc ,sa\_fut ,la\_emph , li\_jus , li\_sub \\
prc2 & 0, wa\_conj \\
prc3 & 0, >a\_ques \\
enc0 & 0, 1p\_dobj, 1s\_dobj, 3ms\_dobj, 2fs\_dobj \\
enc1 & 0, 1p\_dobj, 1s\_dobj, 3ms\_dobj, 2fs\_dobj \\
\bottomrule
\end{tabular}

\vspace{0.8em}

\begin{tabular}{p{0.2\linewidth} p{0.68\linewidth}}
\toprule
\textbf{Noun/Adj  } & \textbf{ } \\
\textbf{ Clitic} & \textbf{Values} \\

\midrule
prc0 & 0, Al\_det \\
prc1 & 0, bi\_prep, li\_prep \\
prc2 & 0,  wa\_conj\\
prc3 & 0, >a\_ques \\
enc0 & 0, 1p\_poss, 1s\_poss, 3ms\_poss, 2fs\_poss \\
\bottomrule
\end{tabular}

\caption[Selected clitic features and values]{
Selected clitic features and their possible values for verbs, nouns,
and adjectives. The value labels are glossed as follows:
\texttt{0} = no clitic;
\texttt{la\_rc} = response-conditional \textit{la};
\texttt{sa\_fut} = future marker \textit{sa};
\texttt{la\_emph} = emphatic particle \textit{la};
\texttt{li\_jus} = jussive \textit{li};
\texttt{li\_sub} = subjunctive \textit{li};
\texttt{wa\_conj} = conjunction \textit{wa};
\texttt{>a\_ques} = interrogative particle \texttt{>a};
\texttt{Al\_det} = determiner \textit{Al};
\texttt{bi\_prep} = preposition \textit{bi}; and
\texttt{li\_prep} = preposition \textit{li}.
In the pronominal labels, \texttt{1}, \texttt{2}, and \texttt{3}
denote person; \texttt{s} and \texttt{p} denote singular and plural;
\texttt{m} and \texttt{f} denote masculine and feminine; and
\texttt{dobj} and \texttt{poss} denote direct object and possessive,
respectively. Thus, for example, \texttt{3ms\_dobj} denotes a
third-person masculine singular direct object.
}
\label{tab:clitics}
\end{table}

\paragraph{Compatibility Restrictions}

The selected clitic values are further constrained by the
morphological features of the baseword. Nominal and adjectival
restrictions involve \textit{case} and \textit{state}, whereas verbal
restrictions involve the verbal group and, for imperfective verbs,
\textit{mood}. Tables~\ref{tab:nominal-clitic-compatibility}
and~\ref{tab:verbal-clitic-compatibility} summarize these restrictions.

\begin{table}[t]
\centering
\footnotesize
\setlength{\tabcolsep}{3pt}
\begin{tabularx}{\columnwidth}{@{}llX@{}}
\toprule
\textbf{Group} & \textbf{Clitic value} & \textbf{Restriction} \\
\midrule

Noun/Adj &
$\texttt{prc1}=\texttt{0}$ &
$\texttt{cas}=*$ \\

Noun/Adj &
$\texttt{prc1}=\texttt{bi\_prep}$ &
$\texttt{cas}=\texttt{g}$ \\

Noun/Adj &
$\texttt{prc1}=\texttt{li\_prep}$ &
$\texttt{cas}=\texttt{g}$ \\

\midrule

Noun &
$\texttt{prc0}=\texttt{0}$ &
No additional state restriction \\

Noun &
$\texttt{prc0}=\texttt{Al\_det}$ &
$\texttt{stt}=\texttt{d}$ \\

\midrule

Adj &
$\texttt{prc0}=\texttt{0}$ &
No additional state restriction \\

Adj &
$\texttt{prc0}=\texttt{Al\_det}$ &
$\texttt{stt}\in\{\texttt{d},\texttt{c}\}$ \\

\midrule

Noun/Adj &
$\texttt{enc0}=\texttt{0}$ &
$\texttt{stt}=*$ \\

Noun/Adj &
$\texttt{enc0}\neq\texttt{0}$ &
$\texttt{stt}=\texttt{c}$ \\

Noun/Adj &
$\texttt{enc0}\neq\texttt{0}$ &
Cannot co-occur with
$\texttt{prc0}=\texttt{Al\_det}$ \\

\bottomrule
\end{tabularx}
\caption{Case and state based compatibility restrictions for
nominal and adjectival clitic configurations. The symbol $*$
indicates that the feature is unrestricted.}
\label{tab:nominal-clitic-compatibility}
\end{table}

\begin{table}[t]
\centering
\footnotesize
\setlength{\tabcolsep}{3pt}
\begin{tabularx}{\columnwidth}{@{}llX@{}}
\toprule
\textbf{Group} & \textbf{Condition} & \textbf{Allowed value} \\
\midrule

\texttt{Active\_IV} &
$\texttt{mod}=*$ &
$\texttt{prc1}=\texttt{0}$ \\

\texttt{Active\_IV} &
$\texttt{mod}\in\{\texttt{e},\texttt{x}\}$ &
$\texttt{prc1}=\texttt{la\_emph}$ \\

\texttt{Active\_IV} &
$\texttt{mod}=\texttt{j}$ &
$\texttt{prc1}=\texttt{li\_jus}$ \\

\texttt{Active\_IV} &
$\texttt{mod}=\texttt{s}$ &
$\texttt{prc1}=\texttt{li\_sub}$ \\

\texttt{Active\_IV} &
$\texttt{mod}=\texttt{i}$ &
$\texttt{prc1}=\texttt{sa\_fut}$ \\

\midrule

\texttt{Passive\_IV} &
$\texttt{mod}=*$ &
$\texttt{prc1}=\texttt{0}$ \\

\texttt{Passive\_IV} &
$\texttt{mod}\in\{\texttt{e},\texttt{x}\}$ &
$\texttt{prc1}=\texttt{la\_emph}$ \\

\texttt{Passive\_IV} &
$\texttt{mod}=\texttt{j}$ &
$\texttt{prc1}=\texttt{li\_jus}$ \\

\texttt{Passive\_IV} &
$\texttt{mod}=\texttt{s}$ &
$\texttt{prc1}=\texttt{li\_sub}$ \\

\texttt{Passive\_IV} &
$\texttt{mod}=\texttt{i}$ &
$\texttt{prc1}=\texttt{sa\_fut}$ \\

\midrule

\texttt{Active\_PV} &
-- &
$\texttt{prc1}\in
 \{\texttt{0},\texttt{la\_rc}\}$ \\

\texttt{Passive\_PV} &
-- &
$\texttt{prc1}\in
 \{\texttt{0},\texttt{la\_rc}\}$ \\

\midrule

\texttt{CV} &
-- &
$\texttt{prc1}=\texttt{0}$ \\

\texttt{CV} &
-- &
$\texttt{prc3}=\texttt{0}$ \\

\bottomrule
\end{tabularx}
\caption{Verbal-group compatibility restrictions for the selected
proclitic configurations. Imperfective \texttt{prc1} values are
conditioned by mood (\texttt{mod}); command verbs (\texttt{CV}) do
not admit a \texttt{prc3} proclitic. The symbol $*$ denotes the
underspecified mood value.}
\label{tab:verbal-clitic-compatibility}
\end{table}

\section{Model Configuration and Time \& Cost Analysis}

All models were evaluated with a temperature of 0 and a
maximum output length of 256 tokens. The open-weight models were evaluated locally using NVIDIA A100 and V100 GPUs.

Table~\ref{tab:time_cost_by_model} reports the aggregate runtime and
monetary cost of the complete zero-shot and 10-shot experiments.
GPT-5.4 achieves the strongest overall performance among the evaluated
LLMs and incurs the highest reported monetary cost, totaling
\$1,996.60. Gemini-3.1-Flash-Lite incurs a total cost of \$1,047.10.
The open-weight models have no API usage cost; however, the
computational and infrastructure costs associated with running these
models locally are not included.

GPT-5.4 records the lowest aggregate runtime among the evaluated LLMs,
requiring 8,104 minutes. Reference generation using the CAMeL Tools generator with the CamelMorph MSA database required 233 
minutes without an API cost, making it substantially faster than all
LLM-based systems. These aggregate runtimes reflect the model-specific
configurations evaluated in this study, including the applicable
prompting settings and prompt languages, and therefore should not be
interpreted as normalized per-instance latency comparisons. Overall,
GPT-5.4 provides the strongest LLM performance, whereas  CamelMorph
requires the least runtime and reported monetary cost.

\begin{table}[t]
\centering
\small
\setlength{\tabcolsep}{5pt}

\begin{tabular*}{\columnwidth}{
    @{\extracolsep{\fill}}
    l
    r
    r
}
\toprule
\textbf{Model}
& \multicolumn{1}{c}{\textbf{Time (min)}}
& \multicolumn{1}{c}{\textbf{Cost (\$)}} \\
\midrule

GPT-5.4
& 8,104
& 1,996.60 \\

Gemini-3.1-Flash-Lite
& 32,542
& 1,047.10 \\

Fanar-2-27B-Instruct
& 29,392
& 0.00 \\

Jais-2-8B-Chat
& 30,302
& 0.00 \\

Jais-2-70B-Chat
& 44,331
& 0.00 \\

Qwen2.5-7B-Instruct
& 36,322
& 0.00 \\

ALLaM-7B-Instruct-preview
& 40,224
& 0.00 \\

 CamelMorph
& 233
& 0.00 \\

\bottomrule
\end{tabular*}

\caption{Aggregate runtime and reported monetary cost across all
evaluated systems for the complete zero-shot and 10-shot experiments. Cost is reported in US dollars.}
\label{tab:time_cost_by_model}
 \vspace{-12pt}
\end{table}






\begin{table*}[t]
\centering
\small
\setlength{\tabcolsep}{4pt}

\begin{tabular}{l*{12}{r}}
\toprule
& \multicolumn{6}{c}{\textbf{10-Shot}}
& \multicolumn{6}{c}{\textbf{Zero-Shot}} \\
\cmidrule(lr){2-7}
\cmidrule(lr){8-13}

& \multicolumn{3}{c}{\textbf{Precision}}
& \multicolumn{3}{c}{\textbf{Recall}}
& \multicolumn{3}{c}{\textbf{Precision}}
& \multicolumn{3}{c}{\textbf{Recall}} \\
\cmidrule(lr){2-4}
\cmidrule(lr){5-7}
\cmidrule(lr){8-10}
\cmidrule(lr){11-13}

\textbf{Model}
& \multicolumn{1}{c}{\textbf{Diac}}
& \multicolumn{1}{c}{\textbf{Undiac}}
& \multicolumn{1}{c}{\textbf{Norm}}
& \multicolumn{1}{c}{\textbf{Diac}}
& \multicolumn{1}{c}{\textbf{Undiac}}
& \multicolumn{1}{c}{\textbf{Norm}}
& \multicolumn{1}{c}{\textbf{Diac}}
& \multicolumn{1}{c}{\textbf{Undiac}}
& \multicolumn{1}{c}{\textbf{Norm}}
& \multicolumn{1}{c}{\textbf{Diac}}
& \multicolumn{1}{c}{\textbf{Undiac}}
& \multicolumn{1}{c}{\textbf{Norm}} \\
\midrule

\textbf{GPT}$_{en}$
& \textbf{52.6} & \textbf{68.2} & \textbf{69.0}
& \textbf{56.3} & \textbf{73.0} & \textbf{73.8}
& \textbf{48.8} & \textbf{65.6} & \textbf{66.8}
& \textbf{49.7} & \textbf{66.5} & \textbf{67.7} \\

\textbf{Gemini}$_{en}$
& 52.2 & 63.8 & 64.5
& 52.7 & 64.6 & 65.3
& 41.8 & 57.3 & 58.0
& 41.2 & 56.7 & 57.4 \\

\textbf{Fanar}$_{en}$
& 21.4 & 36.5 & 37.1
& 23.7 & 40.4 & 41.2
& 5.0 & 19.3 & 19.8
& 5.6 & 21.4 & 22.0 \\

\textbf{Fanar}$_{ar}$
& 15.8 & 26.8 & 27.1
& 17.4 & 29.7 & 30.1
& 2.6 & 15.3 & 15.7
& 2.9 & 17.1 & 17.5 \\

\textbf{Jais-70B}$_{en}$
& 9.0 & 20.6 & 20.9
& 10.1 & 23.1 & 23.5
& 8.5 & 16.0 & 16.4
& 9.0 & 17.0 & 17.4 \\

\textbf{Jais-70B}$_{ar}$
& 8.1 & 17.4 & 17.7
& 9.1 & 19.4 & 19.7
& 3.9 & 11.0 & 11.7
& 4.2 & 11.8 & 12.6 \\

\textbf{ALLaM}$_{en}$
& 4.4 & 10.1 & 10.3
& 5.2 & 11.9 & 12.1
& 2.8 & 5.9 & 6.0
& 4.7 & 9.2 & 9.4 \\

\textbf{ALLaM}$_{ar}$
& 3.5 & 8.0 & 8.2
& 5.0 & 11.1 & 11.4
& 2.4 & 7.2 & 7.4
& 3.4 & 10.0 & 10.2 \\

\textbf{Qwen}$_{en}$
& 5.0 & 14.5 & 15.1
& 4.9 & 14.3 & 14.9
& 2.0 & 8.9 & 9.3
& 2.2 & 9.6 & 10.0 \\

\textbf{Jais-8B}$_{en}$
& 3.0 & 8.8 & 9.0
& 3.8 & 11.2 & 11.3
& 1.9 & 4.8 & 5.0
& 2.9 & 7.1 & 7.4 \\

\textbf{Jais-8B}$_{ar}$
& 1.8 & 6.5 & 6.8
& 2.5 & 9.0 & 9.4
& 1.7 & 5.2 & 5.3
& 2.9 & 8.5 & 8.7 \\

\bottomrule
\end{tabular}

\caption{Precision and recall under the 10-shot and
zero-shot prompting settings. Bolded values are the largest in each column.}
\label{tab:micro-precision-recall}
\end{table*}

\section{Precision and Recall Results}
\label{app:precision_recall}

Table~\ref{tab:micro-precision-recall} reports the micro-averaged
Precision and Recall scores corresponding to the F1 results presented
in Table~\ref{tab:overall-results}.

\section{Output Cardinality Analysis}

Table~\ref{tab:output-cardinality} reports how often each model produces no answer, exactly one answer, or multiple answers. Most models predominantly produce one answer, while ALLaM and Jais-8B generate multiple answers more frequently, particularly in the zero-shot setting. This analysis describes output behavior rather than answer correctness.

\begin{table*}[th!]
\centering
\small
\setlength{\tabcolsep}{7pt}

\begin{tabular}{l*{6}{r}}
\toprule
& \multicolumn{3}{c}{\textbf{10-Shot}}
& \multicolumn{3}{c}{\textbf{Zero-Shot}} \\
\cmidrule(lr){2-4}
\cmidrule(lr){5-7}

\textbf{Model}
& \multicolumn{1}{c}{\textbf{No Answer}}
& \multicolumn{1}{c}{\textbf{Exactly One}}
& \multicolumn{1}{c}{\textbf{Multiple}}
& \multicolumn{1}{c}{\textbf{No Answer}}
& \multicolumn{1}{c}{\textbf{Exactly One}}
& \multicolumn{1}{c}{\textbf{Multiple}} \\
\midrule

\textbf{Reference}
& 12.9 & 84.0 & 3.2
& 12.9 & 84.0 & 3.2 \\

\midrule

\textbf{GPT}$_{en}$
& 4.7 & 93.3 & 2.0
& 9.9 & 87.6 & 2.4 \\

\textbf{Gemini}$_{en}$
& 8.5 & 91.4 & 0.0
& 10.6 & 89.3 & 0.0 \\

\textbf{Fanar}$_{en}$
& 0.0 & 99.8 & 0.1
& 0.0 & 99.5 & 0.5 \\

\textbf{Fanar}$_{ar}$
& 0.0 & 99.8 & 0.2
& 0.5 & 97.7 & 1.8 \\

\textbf{Jais-70B}$_{en}$
& 0.0 & 98.9 & 1.1
& 5.1 & 94.3 & 0.6 \\

\textbf{Jais-70B}$_{ar}$
& 2.1 & 95.7 & 2.2
& 4.2 & 94.8 & 1.0 \\

\textbf{ALLaM}$_{en}$
& 0.0 & 93.5 & 6.5
& 9.0 & 45.4 & 45.6 \\

\textbf{ALLaM}$_{ar}$
& 0.0 & 87.2 & 12.8
& 0.2 & 78.8 & 21.1 \\

\textbf{Qwen}$_{en}$
& 11.2 & 88.5 & 0.2
& 6.8 & 87.8 & 5.4 \\

\textbf{Jais-8B}$_{en}$
& 0.0 & 72.8 & 27.2
& 0.6 & 60.4 & 38.9 \\

\textbf{Jais-8B}$_{ar}$
& 1.4 & 81.0 & 17.6
& 1.1 & 48.1 & 50.8 \\

\bottomrule
\end{tabular}

\caption{Distribution of output cardinality under the 10-shot and
zero-shot prompting settings. \textit{No Answer} denotes an empty or
Null output, \textit{Exactly One} denotes an output containing one
answer, and \textit{Multiple} denotes an output containing more than
one answer. The reference row reports the cardinality distribution of
the gold answers. Values are percentages and may not sum to exactly
100\% because of rounding. Subscripts \textit{ar} and \textit{en}
indicate the prompt language.}
\label{tab:output-cardinality}
\end{table*}

\section{Prompts}\label{app:prompts}

We use two generalized prompt templates: one in English and one in
Arabic. The templates are instantiated dynamically according to the
part-of-speech, the presence of clitics, and the prompting setting.
Text enclosed in braces denotes a placeholder replaced before the
prompt is sent to the model.

The 10 in-context examples were randomly selected from a pool
disjoint from the benchmark evaluation instances. None of the
selected examples appears in the evaluated benchmark set.

The fully instantiated prompts used in the experiments,
including the POS-specific rules, clitic rules, and 10-shot
examples, are available in the YallaMorph repository
(see Footnote~\ref{fn:yallamorph-repo}).

The placeholder \texttt{\{MORPHOLOGY\_TYPE\}} is replaced with
\textit{Adjectival}, \textit{Nominal}, or \textit{Verbal}, while
\texttt{\{WORD\_TYPE\}} is replaced with \textit{adjective},
\textit{noun}, or \textit{verb}. The class-specific placeholders
contain the features and interpretation rules associated with the
selected part-of-speech. The clitic placeholders contain the
applicable proclitic and enclitic inventories and their ordering
rules; they are replaced with empty strings for baseword generation.
The few-shot placeholder contains the relevant demonstrations in the
few-shot setting and is empty in the zero-shot setting.


\subsection{English Prompt}
\label{app:english-prompt}

\begin{tcolorbox}[
    enhanced,
    breakable,
    colback=gray!3,
    colframe=black!55,
    boxrule=0.6pt,
    arc=1mm,
    left=3mm,
    right=3mm,
    top=2mm,
    bottom=0mm,
    before skip=6pt,
    after skip=6pt,
    title={English Prompt Template},
    fonttitle=\bfseries
]
\small

You are an Arabic
\texttt{\{MORPHOLOGY\_TYPE\}}
Morphological Form Generator.

\medskip
\noindent\textbf{Task}

Given an Arabic
\texttt{\{WORD\_TYPE\}}
lemma, POS, gloss, and a specific set of morphosyntactic
features, generate the corresponding fully diacritized Arabic
\texttt{\{WORD\_TYPE\}}
form or forms.

\medskip
\noindent\textbf{Requirements}

\begin{itemize}[leftmargin=*]
    \item Focus only on Arabic
          \texttt{\{MORPHOLOGY\_TYPE\}}
          morphology.
    \item Output only the fully diacritized Arabic form or forms
          corresponding to the requested feature bundle.
    \item A feature bundle may yield no valid form, exactly one valid
          form, or multiple valid forms for the same slot.
    \item If the feature combination is not realizable for the lemma
          in Arabic, return an \texttt{IMPOSSIBLE} result.
    \item Do not provide explanations, glosses, comments, or reasoning.
    \item Do not invent forms.
    \item If multiple valid forms exist, return all valid forms.
\end{itemize}

\medskip
\noindent\textbf{Input Features}

The input contains the lemma, POS, gloss, gender, and number, followed
by the features associated with the selected part-of-speech.

\medskip
\noindent
\texttt{\{CLASS\_SPECIFIC\_FEATURES\}}

\medskip
\noindent
\texttt{\{CLITIC\_FEATURES\}}

\medskip
\noindent\textbf{Critical Interpretation Rules}

\medskip
\noindent
\texttt{\{CLASS\_SPECIFIC\_RULES\}}

\medskip
\noindent
\texttt{\{CLITIC\_RULES\}}

\medskip

Generate the form strictly from the complete feature bundle. If the
combination of grammatical features or clitics is not morphologically
realizable, return \texttt{IMPOSSIBLE}.

\medskip
\noindent\textbf{Output Format}

For one or more valid forms:

\begin{center}
\texttt{\{"arabic\_forms": ["form1", "form2"], "status": "OK"\}}
\end{center}

If the requested form is impossible:

\begin{center}
\texttt{\{"arabic\_forms": null, "status": "IMPOSSIBLE"\}}
\end{center}

\medskip
\noindent
\texttt{\{N.FEW\_SHOT\_EXAMPLES\}}

\end{tcolorbox}


\subsection{Arabic Prompt}
\label{app:arabic-prompt}

\begin{tcolorbox}[
    enhanced,
    breakable,
    colback=gray!3,
    colframe=black!55,
    boxrule=0.6pt,
    arc=1mm,
    left=3mm,
    right=3mm,
    top=2mm,
    bottom=0mm,
    before skip=6pt,
    after skip=6pt,
    title={Arabic Prompt Template},
    fonttitle=\bfseries
]
\small

\begin{RLtext}
أنت مولّد للصيغ الصرفية ل
\LR{\texttt{\{MORPHOLOGY\_TYPE\_AR\}}} العربيه.

\end{RLtext}

\begin{RLtext}
\noindent\textbf{المهمة}
\end{RLtext}

\begin{RLtext}
عند إعطاء مدخل معجمي
\LR{\texttt{\{WORD\_TYPE\_AR\}}},
وقسم الكلام، والمعنى بالإنجليزية، ومجموعة محددة من السمات
الصرفية، قم بتوليد الصيغة أو الصيغ العربية المشكّلة بالكامل
المقابلة
\LR{\texttt{\{TARGET\_TYPE\_AR\}}}.
\end{RLtext}

\medskip

\begin{RLtext}
\noindent\textbf{المتطلبات}
\end{RLtext}

    \begin{RLtext}
    ركّز فقط على تصريف
    \LR{\texttt{\{MORPHOLOGY\_DOMAIN\_AR\}}}
    العربية.
    \end{RLtext}

    \begin{RLtext}
    أخرج فقط الصيغة أو الصيغ العربية المشكّلة بالكامل التي
    تطابق حزمة السمات المطلوبة.
    \end{RLtext}

    \begin{RLtext}
    قد تؤدي حزمة السمات إلى عدم وجود صيغة صحيحة، أو وجود صيغة
    صحيحة واحدة، أو وجود أكثر من صيغة صحيحة للخانة نفسها.
    \end{RLtext}

    \begin{RLtext}
    إذا كانت تركيبة السمات غير قابلة للتحقق صرفيًا لهذا
    المدخل المعجمي، فأرجع نتيجة
    \LR{\texttt{IMPOSSIBLE}}.
    \end{RLtext}

    \begin{RLtext}
    لا تقدّم شروحات أو معاني أو تعليقات أو خطوات تفكير.
    \end{RLtext}

    \begin{RLtext}
    لا تخترع صيغًا.
    \end{RLtext}

    \begin{RLtext}
    إذا وُجدت عدة صيغ صحيحة، فأرجع جميع الصيغ الصحيحة.
    \end{RLtext}

\medskip

\begin{RLtext}
\noindent\textbf{سمات الإدخال}
\end{RLtext}

\begin{RLtext}
يحتوي الإدخال على المدخل المعجمي، وقسم الكلام، والمعنى
بالإنجليزية، والجنس، والعدد، ثم السمات المرتبطة بقسم الكلام
المحدد.
\end{RLtext}

\medskip

\begin{RLtext}
\LR{\texttt{\{CLASS\_SPECIFIC\_FEATURES\_AR\}}}
\end{RLtext}

\medskip

\begin{RLtext}
\LR{\texttt{\{CLITIC\_FEATURES\_AR\}}}
\end{RLtext}

\medskip

\begin{RLtext}
\noindent\textbf{قواعد تفسيرية مهمة}
\end{RLtext}

\medskip

\begin{RLtext}
\LR{\texttt{\{CLASS\_SPECIFIC\_RULES\_AR\}}}
\end{RLtext}

\medskip

\begin{RLtext}
\LR{\texttt{\{CLITIC\_RULES\_AR\}}}
\end{RLtext}

\medskip

\begin{RLtext}
ولّد الصيغة بدقة بناءً على حزمة السمات الكاملة. إذا كانت
تركيبة السمات النحوية أو السوابق أو اللواحق غير قابلة للتحقق
صرفيًا، فأرجع
\LR{\texttt{IMPOSSIBLE}}.
\end{RLtext}

\medskip



\begin{RLtext}
\noindent\textbf{تنسيق الإخراج}
\end{RLtext}

\begin{RLtext}
عند وجود صيغة صحيحة واحدة أو أكثر:
\end{RLtext}

\begin{center}
\texttt{\{"arabic\_forms": ["form1", "form2"], "status": "OK"\}}
\end{center}

\begin{RLtext}
إذا كانت الصيغة المطلوبة غير قابلة للتحقق:
\end{RLtext}

\begin{center}
\texttt{\{"arabic\_forms": null, "status": "IMPOSSIBLE"\}}
\end{center}

\medskip

\begin{RLtext}
\LR{\texttt{\{N.FEW\_SHOT\_EXAMPLES\_AR\}}}
\end{RLtext}

\end{tcolorbox}

\paragraph{Example Instantiation.}
The following example illustrates how the feature placeholders in the
English and Arabic template are instantiated for a verbal item. It requests the
third-person masculine singular imperfective indicative active form of the lemma \<كَتَب>.

\begin{tcolorbox}[
    enhanced,
    colback=gray!3,
    colframe=black!55,
    boxrule=0.6pt,
    arc=1mm,
    left=3mm,
    right=3mm,
    top=2mm,
    bottom=0mm,
    before skip=6pt,
    after skip=6pt,
    title={Baseword Example of the English Prompt},
    fonttitle=\bfseries
]
\small

\noindent\textbf{Input to process}

\medskip

\noindent
\textbf{LEMMA:} \<كَتَب>

\noindent
\textbf{POS:} Verb

\noindent
\textbf{GLOSS:} write

\noindent
\textbf{GENDER:} Masculine

\noindent
\textbf{NUMBER:} Singular

\noindent
\textbf{PERSON:} 3rd

\noindent
\textbf{ASPECT:} Imperfective

\noindent
\textbf{MOOD:} Indicative

\noindent
\textbf{VOICE:} Active

\end{tcolorbox}

\begin{tcolorbox}[
    enhanced,
    colback=gray!3,
    colframe=black!55,
    boxrule=0.6pt,
    arc=1mm,
    left=3mm,
    right=3mm,
    top=2mm,
    bottom=0mm,
    before skip=6pt,
    after skip=6pt,
    title={Baseword Example of the Arabic Prompt},
    fonttitle=\bfseries
]
\small

\begin{RLtext}
\noindent\textbf{المدخل المطلوب معالجته}

\medskip

\noindent
\textbf{المدخل المعجمي:} كَتَب

\noindent
\textbf{قسم الكلام:} فعل

\noindent
\textbf{المعنى بالإنجليزية:}
\LR{\texttt{write}}

\noindent
\textbf{الجنس:} مذكر

\noindent
\textbf{العدد:} مفرد

\noindent
\textbf{الشخص:} الغائب

\noindent
\textbf{الزمن:} مضارع

\noindent
\textbf{العلامة الإعرابية:} المرفوع

\noindent
\textbf{البناء:} مبني للمعلوم
\end{RLtext}

\end{tcolorbox}

\newpage
\section{Licenses}
\label{app:license}

We list below the licenses of the data and tools used in this work, all of which are employed in accordance with their intended use.

\begin{itemize}[
  itemsep=1pt,
  parsep=1pt,
  topsep=2pt,
  partopsep=0pt
]
    \item BAREC-10M Corpus~\cite{elmadani-etal-2026-large}: Creative Commons Attribution Share Alike 4.0
    
    \item CAMeLBERT Frequency List~\cite{Khalifa:2021:Camel_Frequency}: Creative Commons Attribution Share Alike 4.0

    \item CAMeL Tools~\citep{obeid-etal-2020-camel} and  CamelMorph ~\citep{khairallah-etal-2024-camel} : MIT License
    
    \item CamelMorph MSA morphological database,
lexicon, and specification data
\citep{khairallah-etal-2024-camel}:
Creative Commons Attribution 4.0 International
(CC BY 4.0).
    \item ALLaM-7B-Instruct ~\cite{allam}: Apache License 2.0
    \item Fanar-2-27B-Instruct ~\cite{fanar2}: Apache license 2.0
    \item Jais-2-8B-Chat~\cite{jais2}: Apache license 2.0
    \item Jais-2-70B-Chat~\cite{jais2}: Apache License 2.0

    \item Qwen2.5-7B-Instruct~\cite{qwen2.5}: Apache License 2.0

\end{itemize}

\end{document}